\documentclass{article}
\PassOptionsToPackage{sort&compress}{natbib}
\usepackage[final]{corl_2026}
\usepackage{graphicx}
\usepackage[font=small,labelfont=bf,skip=4pt]{caption}
\IfFileExists{booktabs.sty}{\usepackage{booktabs}}{%
  \newcommand{\toprule}{\hline}\newcommand{\midrule}{\hline}\newcommand{\bottomrule}{\hline}}
\usepackage{amsmath}

\makeatletter
\renewcommand{\section}{\@startsection{section}{1}{\z@}%
  {-0.9ex \@plus -0.2ex \@minus -0.2ex}{0.4ex \@plus 0.2ex}{\large\bf\raggedright}}
\g@addto@macro\normalsize{%
  \setlength\abovedisplayskip{2pt plus 1pt minus 1pt}\setlength\belowdisplayskip{2pt plus 1pt minus 1pt}%
  \setlength\abovedisplayshortskip{1pt plus 1pt}\setlength\belowdisplayshortskip{2pt plus 1pt}}
\renewcommand{\@maketitle}{%
  \vbox{%
    \hsize\textwidth
    \linewidth\hsize
    \vskip 0.1in
    \centering
    {\LARGE\bf \@title\par}
    \begin{tabular}[t]{c}\bf\rule{\z@}{14\p@}\@author\end{tabular}%
    \vskip 0.12in \@minus 0.04in
  }
}
\makeatother

\newif\ifcomments\commentsfalse
\definecolor{frankc}{rgb}{0.0,0.35,0.75}
\definecolor{haonanc}{rgb}{0.7,0.35,0.0}
\definecolor{pengfeic}{rgb}{0.8,0.0,0.0}
\ifcomments
  \newcommand{\frank}[1]{\unskip\space{\color{frankc}[Frank: #1]}}
  \newcommand{\haonan}[1]{\unskip\space{\color{haonanc}[Haonan: #1]}}
  \newcommand{\pengfei}[1]{\unskip\space{\color{pengfeic}[Pengfei: #1]}}
\else
  \newcommand{\frank}[1]{}\newcommand{\haonan}[1]{}\newcommand{\pengfei}[1]{}
\fi

\title{Reins: Frontier Multimodal Models Driving Real Robot Arms on Long-Horizon Tasks, Zero-Shot}

\author{
  Pengfei Ye$^{1}$\qquad Zhiyang Dou$^{1}$\qquad Yuxiang Ma$^{1}$\\
  \bf Haonan Chen$^{2}$\qquad Heng Yang$^{2}$\qquad Edward H. Adelson$^{1}$\\
  \normalfont $^{1}$MIT CSAIL\qquad $^{2}$Harvard University\qquad\texttt{pfy712@mit.edu}\\
  \url{https://reins-robot.github.io}
}
\begin{document}
\maketitle
\vspace{-2mm}

\begin{abstract}
Frontier models such as GPT-6.1 Sol, GPT-6 Astra and Claude Opus 5.5 can now plan and call tools, yet they
still struggle to drive real robots through long-horizon, contact-rich manipulation, blind while the arm
moves and while they think. We present Reins, a harness in which an off-the-shelf model operates a
dual-arm robot with 3D perception tools and atomic actions that report contact and grip. The model chains
actions in one call, each with a condition that must hold, stopping at the first that fails; a second
instance, the narrator, watches the cameras continuously and reports what changed with every result.
From images, Reins also builds a simulation-ready twin of the scene for continual policy learning. With
natural-language prompts and no task-specific code, \mbox{GPT-6.1} Sol succeeds in 88\% of real episodes
on five tasks; without the harness, it fails every shell game, taking 8 times as long. Three frontier
models perform alike but differ in tracking a scene that changes while the actor thinks. Code will be
released upon publication.
\end{abstract}

\begin{figure}[h]
  \centering
  \vspace{0mm}
  \includegraphics[width=0.97\linewidth]{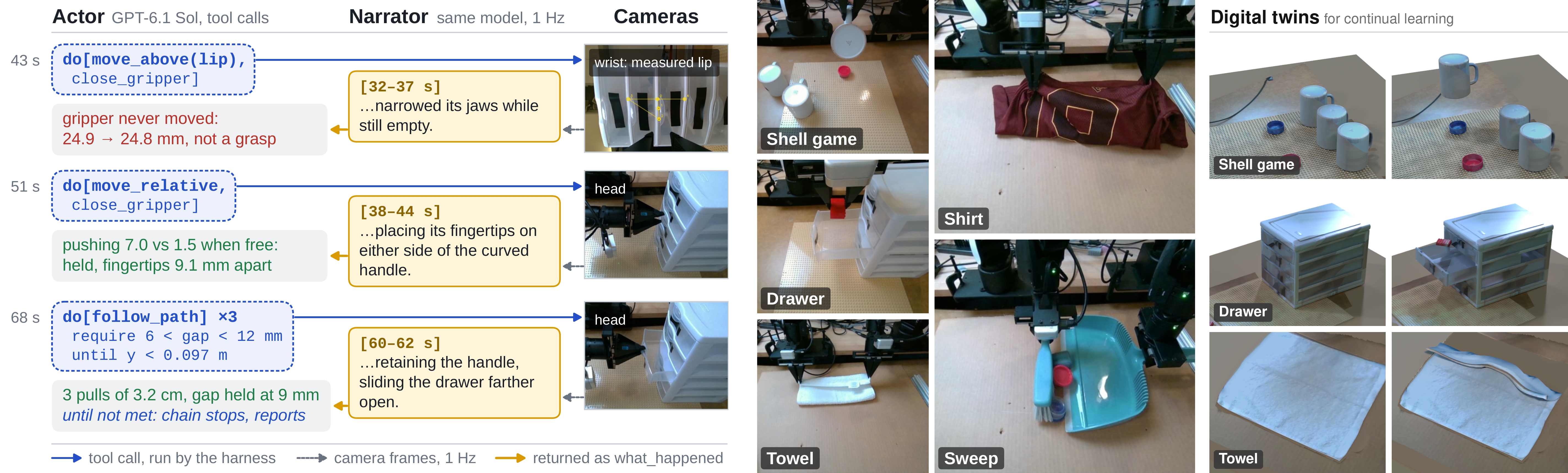}
  \caption{\textbf{An overview of Reins.} Left: three calls of a Sol drawer episode; results report what the robot
  measured, a chain of pulls stops when its condition fails, and the narrator's account arrives with the
  next result. Middle: the five tasks. Right: sim-ready digital twins, each built from one head-camera frame, at rest and running in MuJoCo
  (App.~\ref{sec:twin}), for continual learning; only object dynamics are shown.}
  \label{fig:overview}
  \vspace{-1mm}
\end{figure}

\section{Introduction and Related Work}
Frontier models that plan and call tools~\citep{react2023,toolformer2023} now run as robot policies
without robot training, mostly in simulation, where benchmarks find a gap between understanding and
executing a task~\mbox{\citep{galbot2026,astranomy2026,libero10bench2026,agent2act2026,gptgear2026,demofree2026}}. On real
hardware, Show-Harness~\citep{showharness2026} moves a Franka and a bimanual AgileX in 2--4\,cm steps,
mostly pick-and-place. Trained policies struggle too: on RoboDojo's real \emph{Sweep Blocks}, ten
vision-language-action models trained on 100 demonstrations each never succeed~\citep{robodojo2026}.

Earlier language-model systems compose skills, write programs or spatial targets, or feed scene
descriptions back to the model~\citep{saycan2022,text2motion2023,codeaspolicies2023,progprompt2023,chatgptrobotics2024,voxposer2023,rekep2024,moka2024,innermonologue2022,socratic2023,reflect2023};
others learn control from robot
data~\citep{aloha2023,diffusionpolicy2023,rt22023,openvla2024,pi02025,pi052025,geminirobotics2025} or
sequence learned skills~\citep{geminier2026,roboclaw2026,autort2024}. In Reins, the model drives the
arms through atomic actions, with no skill library or task code; two choices set it apart. First,
the model attaches stopping conditions to each chain of actions, checked at every step, as in
guarded moves~\citep{will1975,lozanoperez1984} and the action conditions of task and motion
planning~\citep{garrett2021tamp}, but written at run time; Show-Harness runs its action
chunks open-loop~\citep{showharness2026}. Second, another instance of the model, the narrator, watches
every camera frame while the actor thinks and keeps a running account; VoLo's agent
monitors its own execution at 0.2\,Hz~\citep{volo2026}. We contribute
(i)~Reins with its guarded chains, (ii)~the narrator and its latency analysis, and (iii)~a real-robot evaluation of three frontier models on five long-horizon tasks, with time and cost.

\begin{figure}[t]
  \centering
  \vspace{-9mm}
  \begin{minipage}[t]{0.45\textwidth}
    \vspace{0pt}\includegraphics[width=\linewidth]{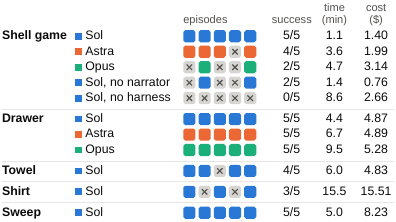}
  \end{minipage}\hfill
  \begin{minipage}[t]{0.45\textwidth}
    \vspace{0pt}\includegraphics[width=\linewidth]{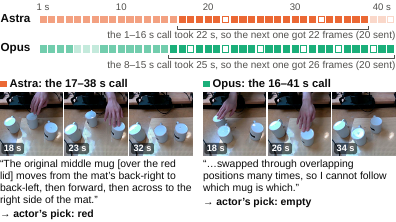}
  \end{minipage}
  \caption{\textbf{Results and the narrator.} (a)~Five episodes each ($\times$: failure); time: median
  minutes; cost: mean API cost. (b)~Astra's and Opus's narrators (standard tier) in a shell
  game: one square per frame, shaded by call.}
  \label{fig:results}
  \vspace{0mm}
\end{figure}

\section{Method}
\textbf{Actor loop.} At decision $k$, the actor, an off-the-shelf model $\pi$ given a task prompt $g$,
outputs a tool call or chain of calls $a_k\sim\pi(\cdot\mid g,h_k)$; its result $y_k=(r_k,I_k,q_k,E_k)$
(measurements, new images, both arms' state and the narrator's entries since $y_{k-1}$) joins the
history $h_{k+1}$ (App.~\ref{app:context}). An episode of $K$ decisions lasts $T=\sum_{k}(\lambda_k+d_k)$, model latency plus
execution. A chain of thought takes time: latency is 41--78\% of $T$ in our runs, so each decision
should cover several checked steps (1.6--3.4 on average), and the scene may change while the actor thinks.
\textbf{Perception and atomic actions.} Of the eighteen tools (App.~\ref{app:tools}), \texttt{find}
(SAM~3~\citep{sam3} on RGB-D) and \texttt{measure} turn an object name or image point into a 3D point in
the base frame, to about 1.6\,mm with the arm's own wrist camera. Atomic actions move an arm, tilt its
tool or work its gripper; \texttt{move\_until\_contact} stops when joint current indicates contact, and
\texttt{close\_gripper} reports whether the fingers hold anything.
\textbf{Guarded chains.} \texttt{do} executes a chain $P=\big((p_i,\rho_i,u_i,n_i)\big)_{i=1}^{n}$ of up
to eight calls and sixteen runs. Step $i$ runs tool $p_i$ up to $n_i$ times until $u_i$ holds, and each
result $r_{i,m}$ must pass $\rho_i$. The chain stops after its last step or at the first run that fails
its check or uses up its repeats:
\begin{equation}
  (i,m)^\star=\min\big\{(i,m)\;:\;\neg\rho_i(r_{i,m})\;\lor\;\big(m=n_i\land\neg u_i(r_{i,m})\big)\big\}.
  \label{eq:program}
\end{equation}

\textbf{Narrator.} A second instance $\nu$ of the model watches the head camera at 1 frame/s in one
ongoing conversation. Call $j$ receives the frames in $W_j=(\tau_{j-1},\tau_j]$ (at most 20, plus wrist
frames while an arm moves) and, after latency $\ell_j$, returns an entry $e_j=\nu(W_j,e_{<j})$ that
reaches the actor as \texttt{what\_happened}. The next window closes when the call returns,
$\tau_{j+1}=\tau_j+\ell_j$, so
\begin{equation}
  |W_{j+1}|=\ell_j,\qquad \ell_j\;\le\; t-\tau_j\;<\;\ell_j+\ell_{j+1}
  \quad\mbox{for } t\in[\tau_{j+1},\tau_{j+2}){:}
  \label{eq:narrator}
\end{equation}
each window lasts as long as the previous call, and the newest frame reported is 1--2 latencies old.

\section{Experiments and Findings}
\textbf{Setup and tasks.} Two 6-DoF AirBot Play arms with stock parallel grippers share a table, with
RGB-D cameras on the head and each wrist; episodes start at the zero pose and the operator judges
success. We run five episodes per condition with GPT-6.1 Sol (Ultrafast tier), GPT-6 Astra and Claude
Opus 5.5, at high reasoning effort. In the \emph{shell game}, the first of three mugs lifted after a
20--30\,s shuffle must cover a hidden red object; the \emph{drawer} is opened by its 3--4\,mm lip,
filled with a block and closed; a \emph{towel} is folded twice, a \emph{shirt} sleeves-in and then twice;
in \emph{sweep}, one arm holds a dustpan and the other brushes objects in. The drawer and sweep prompts add grip and pose hints.

\textbf{The narrator and the harness.} All three models open the drawer every time; the shell game, whose
scene changes while the actor waits, separates them. Without the narrator, Sol wins only 2 of 5. By
Eq.~\eqref{eq:narrator}, a faster narrator sees more: Sol's (Ultrafast tier) answers in 1--6\,s and sees a
shuffle in 8--11 windows; Astra's and Opus's (standard tier) take 3--40\,s and see it in 2--4
(Fig.~\ref{fig:results}b). Astra's narrator still tracks the red object in 4 of 5 episodes, Opus's in
none; Opus's two wins were guesses. Without the harness (raw images, straight-down moves, \texttt{wait},
\texttt{look}), Sol never lifts the correct mug: in four episodes it closes the gripper 6--9 times in 60 calls
but lifts no mug, and once it lifts the wrong one.

\section{Limitations and Future Work}
\textbf{Limitations.} Task knowledge enters only through natural-language prompts; contact is sensed
only from joint current and gripper width; we test one platform, five episodes per condition; and a
slow narrator reports fast changes late (Eq.~\eqref{eq:narrator}). \textbf{Future work.} Every episode
is fully logged (100\,Hz state, images, text); we will turn successful episodes into reusable chains and
checks, rehearsed in simulation twins (App.~\ref{sec:twin}), so that the same model finishes tasks faster, from shorter prompts.

\bibliography{refs}

\clearpage
\appendix

\section{Context, Caching and Memory}
\label{app:context}

\textbf{What the actor keeps.} The text of an episode stays in the history until the end: every
reasoning step, tool result and measurement. Images do not. A result's images remain available for the
following decisions, but once the prompt has grown by 30k tokens since the last compaction, all images
except those of the newest image-bearing turn are replaced at once by one-line stubs. What the model
derived from an old image remains: the object ids that \texttt{find} registered, the pixels it measured
and the results. A chain returns only the images of its last run.

\textbf{Why rarely and in bulk.} On the provider route we use, the prompt cache only reuses a prompt that
extends one it has already served in full, so any edit of earlier history costs one full re-read. Two
alternatives performed worse. Compacting whenever the text exceeded a fixed budget edited the history on
every call: after the first compaction of a 65-call episode, every call re-read 45--54k tokens at full
price, 90\% of that episode's uncached tokens. A sliding window that kept the two newest image turns and
edited the history before every call reduced the cached prefix to the 6.3k-token system prompt and raised
the actor's cost per call from \$0.24 to \$0.85 on the Ultrafast tier, without making calls faster.
Compaction triggered by growth costs one re-read per compaction, after which the cache holds again. For
Claude, whose provider caches only a prefix that the request marks, the harness adds the marker.

\textbf{The narrator's context.} The narrator reports by exception: it replies only with what a window
adds (an entry, a correction, or \texttt{NO CHANGE}), and the harness stamps the reply with the window's
time span and appends it; consecutive quiet windows are merged into one line. An earlier version that
rewrote the whole account on every call spent 11\,s and 666 output tokens on a window in which nothing
had changed. The calls form one continuing conversation, so each window is a new turn on a cached
prefix; re-sending the account in a fresh prompt for each window was cached in none of 41 calls. The
conversation restarts every 60 frames (about a minute), seeded with the episode's first frame and the
account so far, so old frames are dropped in bulk with one uncached read per restart. We count frames
rather than tokens because models encode a frame differently: Claude Opus 5.5 spends about 600--700
tokens on a frame that GPT-6 encodes in about 240, so a 15k-token budget restarted Opus's narrator every
1.9 calls (43 restarts in one drawer episode, with 32\% of its input cached). A 60k-token ceiling still
bounds every model's conversation.

\textbf{Memory within an episode.} Besides the text history and the narrator's account, four records
persist across calls. (i)~\emph{Objects}: \texttt{find} registers each object as an id with a 3D
position, drawn on every later image and listed in every result. (ii)~\emph{Measured shapes}:
\texttt{measure} can turn a visible contour, such as a cloth edge or a drawer lip, into base-frame points
that remain valid after the arm occludes it, and \texttt{follow\_path} can travel through them. (iii)~\emph{A
task ledger}: \texttt{plan} records the steps and \texttt{mark\_step} the evidence for each; \texttt{done}
is refused while a planned step is pending. (iv)~\emph{What each arm holds}, with the finger gap measured
at the grasp. The narrator names identical objects by where it first saw them (``the original middle
mug''), which lets an identity survive a shuffle.

\textbf{Measured.} Table~\ref{tab:context} summarizes the 45 counted episodes.

\begin{table}[!h]
  \centering
  \small
  \begin{tabular}{lcccc}
    \toprule
    & \multicolumn{2}{c}{input tokens from cache} & compactions per episode & largest prompt \\
    model & actor & narrator & median (max) & median (max) \\
    \midrule
    GPT-6.1 Sol     & 90\% & 73\% & 1 (5) & 61k (171k) \\
    GPT-6 Astra     & 91\% & 68\% & 1 (2) & 45k (65k) \\
    Claude Opus 5.5 & 88\% & 56\% & 2 (3) & 71k (114k) \\
    \bottomrule
  \end{tabular}
  \caption{Context and caching over the counted episodes. The narrator's share is lower because each
  restart re-reads its first frame and the account at full price.}
  \label{tab:context}
\end{table}

\clearpage
\section{Tools}
\label{app:tools}

Table~\ref{tab:tools} lists the eighteen tools the actor can call. Every result returns the state of
both arms (tool position and angles, gripper gap, what each arm holds), fresh camera images and the
narrator's entries since the previous result. On the dual-arm robot, the motion and gripper tools take
the name of the arm to move.

\begin{table}[!h]
  \centering
  \small
  \begin{tabular}{@{}llp{0.6\linewidth}@{}}
    \toprule
    group & tool & what it does \\
    \midrule
    perception & \texttt{look} & returns fresh images from the head and wrist cameras, with known object ids drawn on them \\
      & \texttt{find} & detects every instance of a named object in one camera (SAM~3 on RGB-D) and registers each as an id with a 3D position in the base frame \\
      & \texttt{measure} & turns one pixel, or an ordered list of pixels, into base-frame 3D points \\
    \midrule
    motion & \texttt{move\_above} & moves the gripper to an object or point plus offsets, with given tool angles \\
      & \texttt{move\_relative} & moves the gripper by $(dx, dy, dz)$, at most 0.15\,m per axis \\
      & \texttt{descend} & moves straight down to an object's grasp height, to a given height or by a given distance \\
      & \texttt{lift} & moves straight up by a given height \\
      & \texttt{tilt} & rotates the tool in place: pitch tilts the approach away from vertical, roll spins it about its axis \\
      & \texttt{move\_until\_contact} & moves along a straight line and stops when the joint current indicates contact, read about 25 times per second \\
      & \texttt{follow\_path} & moves through a list of points in one continuous motion, holding the tool angles \\
      & \texttt{home} & returns the arm to its home pose with the gripper open \\
      & \texttt{wait} & waits for up to 120\,s while the narrator keeps watching, then observes \\
    \midrule
    gripper & \texttt{open\_gripper} & opens the gripper to a given gap \\
      & \texttt{close\_gripper} & closes in small increments and stops when the fingers stop; reports whether something is held from where the closure stopped \\
    \midrule
    task & \texttt{plan} & records the ordered steps of the task \\
      & \texttt{mark\_step} & updates one step with the evidence from a new observation \\
      & \texttt{done} & ends the episode; refused while a planned step is pending \\
    \midrule
    chain & \texttt{do} & runs up to eight of the tools above, sixteen runs in all, with a check and an optional repeat condition on each step (Eq.~\eqref{eq:program}) \\
    \bottomrule
  \end{tabular}
  \caption{The actor's tools.}
  \label{tab:tools}
\end{table}

\section{Simulation-Ready Twins of the Scene}
\label{sec:twin}

Reins can also rebuild the scene in front of the robot as a MuJoCo simulation from one calibrated
head-camera RGB frame. This tool, \texttt{model\_scene}, was not used in the episodes of Section~3. The
model names the objects in the image and gives each a category, a box and any hint it can read off the
image, such as an upside-down mug or a hidden handle. The tool returns a scene in the robot's world frame
with, for every object, a closed mesh, collision geometry, mass, inertia, joints and surface appearance,
together with the calibrated head camera, the bases of both arms and a verdict on whether each object is
ready for simulation.

\textbf{Geometry.} Each object is built as a parametric model of its category, by code that drives
Blender, rather than fused from depth: a mug is a hollow lathe profile with a swept handle, a drawer
cabinet has its drawers on prismatic joints, and a towel or shirt is a deformable sheet; bricks, caps,
sponges, brushes and dustpans have builders of their own, and a new category is one more builder.
SAM (ViT-H)~\citep{kirillov2023sam} segments each object from its box, and the mask's footprint on the
table gives a first guess of position, axis and size. The logs keep only the head camera's RGB frames, so depth comes from a monocular network, scaled by the table's
calibrated height. The models are then fitted to the masks through the calibrated camera by silhouette
IoU, with Nelder--Mead over position, yaw and size, and discrete structure, such as a brick's stud count
or the side a mug's handle is on, is found by enumeration. Collision geometry is built for contact: a
mug's wall is a ring of 16 thin boxes, so an upturned mug can cover a cap and push it along.

\textbf{Appearance and physics.} Appearance comes from inverse rendering. The scene is rendered in white
from the head camera under a uniform sky and under 33 directional lights, and the frame is explained as
\begin{equation}
  I(\mathbf{x}) \;\approx\; \boldsymbol{\rho}_{m(\mathbf{x})} \odot
  \Big( \mathbf{c} \sum_{k=1}^{33} w_k\, S_k(\mathbf{x}) \;+\; \mathbf{s}\, K(\mathbf{x}) \Big),
  \qquad w_k \ge 0,
  \label{eq:twin-light}
\end{equation}
where $S_k$ and $K$ are the renders under light $k$ and under the sky, $\mathbf{c}$ and $\mathbf{s}$ the
colours of the lights and of the sky, and $\boldsymbol{\rho}_m$ the albedo of material $m$. Alternating
non-negative least squares solves for all of them on the pixels where model and mask agree, with white
balance anchored on a surface known to be white, and the albedo is baked into each object's texture
wherever the camera saw it. Mass and inertia come from the closed meshes and material densities, and the
scene settles in MuJoCo~\citep{todorov2012mujoco} before it is checked.

\textbf{Sim-ready gate.} Each movable body must pass the checks in Table~\ref{tab:twin-gate}. Collision
fidelity is the 95th percentile of the distance from the collision primitives' surface outside the
closed mesh to the mesh, with inside and outside decided by the generalized winding
number~\citep{jacobson2013winding}; agreement with the frame is the 90th percentile of the distance
between the model's outline and the mask's. A cloth, simulated as a MuJoCo flex, is checked at rest.

\begin{table}[!h]
  \centering
  \small
  \begin{tabular}{@{}lll@{}}
    \toprule
    check & measure & pass \\
    \midrule
    closed mesh & non-manifold edges; signed volume & $0$; $>0$ \\
    collision fidelity & primitives' surface outside the mesh (p95) & $\le 3$\,mm \\
    stays put & drift during 1--1.5\,s of simulation & $\le 5$\,mm, $\le 3^\circ$ \\
    matches the frame & outline distance to the SAM mask (p90) & $\le 3$\,px \\
    no initial contact & deepest penetration while settling & $\le 2$\,mm \\
    \midrule
    cloth, at rest & drift over the last 0.25\,s; edge strain (p95); outline distance
      & $\le 1$\,mm; $\le 2\%$; $\le 3$\,px \\
    \bottomrule
  \end{tabular}
  \caption{The sim-ready gate, applied to each movable body.}
  \label{tab:twin-gate}
\end{table}

\begin{figure}[!h]
  \centering
  \includegraphics[width=\linewidth]{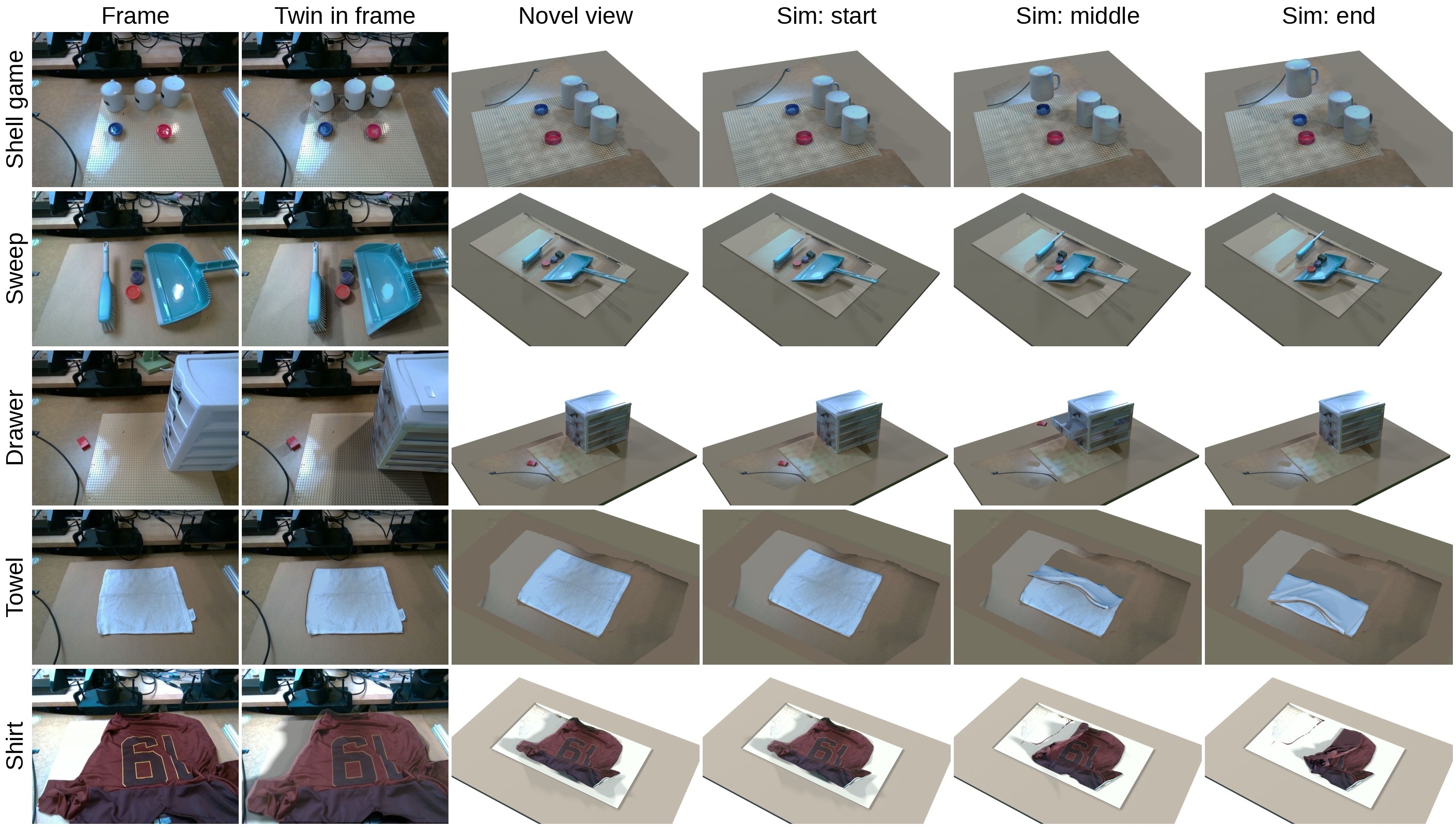}
  \caption{Twins of the five tasks of Section 3, each from one head-camera RGB frame. Columns, from left: the
  frame; the twin rendered into the frame from the same calibrated camera; the twin from another view; and
  the start, middle and end of the scripted task in MuJoCo, rendered in Blender.}
  \label{fig:twin-overview}
\end{figure}

\textbf{Results.} We built a twin from the first frame of one episode of each of the five tasks of Section 3 (Table~\ref{tab:twin-results}, Fig.~\ref{fig:twin-overview}). Of the 14 movable objects, 11 passed the gate: all five in the shell
game, three of five in the sweep, one of the two in the drawer scene, and both cloths. The certified
objects overlap their masks with IoU 0.90--0.99, their outlines lie 0.7--2.4\,px from the models', and
their collision geometry stays within 0.2--2.1\,mm of their surfaces; the recovered lighting reproduces
the frame to within 11--19 of 255 levels. Enumerating stud counts identified the drawer scene's brick as
a $6\times4$ brick and raised its IoU from 0.75 to 0.90, and in the shell game the three mugs, fitted
independently, agree in diameter and in height to within 1.3\,mm. In every twin a scripted task, driven
by mocap hands rather than by the model, runs to completion: a mug covers a cap and pushes it 70\,mm, a
brush sweeps two caps and a sponge into a held dustpan, a drawer opens 120\,mm, takes the brick and closes
without moving the cabinet, and the towel and the shirt are folded. Building a scene of rigid objects
takes 9--12\,min on 12 CPU cores.

\begin{table}[!h]
  \centering
  \small
  \setlength{\tabcolsep}{4pt}
  \begin{tabular}{@{}lccccl@{}}
    \toprule
    task & certified & IoU & outline (px) & light & scripted task in MuJoCo \\
    \midrule
    shell game & 5/5 & 0.92--0.96 & 0.95--1.64 & 11.4 & cover a cap, push it 70\,mm \\
    sweep & 3/5 & 0.95--0.97 & 0.68--1.37 & 17.6 & sweep 2 caps and a sponge into the pan \\
    drawer & 1/2 & 0.90 & 1.37 & 19.0 & open 120\,mm, place the brick, close \\
    towel & 1/1 & 0.99 & 1.16 & 16.2 & fold in half (296 $\to$ 151\,mm) \\
    shirt & 1/1 & 0.99 & 2.39 & 17.8 & fold over (678 $\to$ 441\,mm) \\
    \bottomrule
  \end{tabular}
  \caption{Twins of the five tasks of Section 3, each built from one head-camera RGB frame. Certified: objects that pass
  the gate (a cabinet and its drawers count as one). IoU and outline distance are over the certified
  objects; light is the mean absolute error of Eq.~\eqref{eq:twin-light} in 8-bit sRGB levels.}
  \label{tab:twin-results}
\end{table}

\end{document}